\documentclass[a4paper,fleqn]{cas-dc}
\usepackage[numbers]{natbib}
\usepackage{array}
\usepackage{textcomp}
\usepackage{stfloats}
\usepackage{url}
\usepackage{verbatim}
\usepackage{graphicx}
\usepackage{amsmath,amssymb,amsfonts}
\usepackage{graphicx}
\usepackage{textcomp}
\usepackage{xcolor}
\usepackage{colortbl}
\usepackage{multirow}
\usepackage{bm}
\usepackage{stfloats}
\usepackage{float}
\usepackage{algorithm}
\usepackage{tabularray}
\usepackage{algorithmicx}  
\usepackage{algpseudocode}  
\usepackage{setspace}
\usepackage{balance}

\def\tsc#1{\csdef{#1}{\textsc{\lowercase{#1}}\xspace}}
\tsc{WGM}
\tsc{QE}

\begin{document}
\let\WriteBookmarks\relax
\def\floatpagepagefraction{1}
\def\textpagefraction{.001}
\let\printorcid\relax 

\shorttitle{}    

\shortauthors{Haochen Shi et al.}

\title[mode = title]{A Pre-trained Data Deduplication Model based on Active Learning}  

\author[1,2]{Haochen Shi}
\author[1]{Xinyao Liu}
\author[1]{Fengmao Lv}
\author[1]{Hongtao Xue}
\author[1]{Jie Hu}
\author[1,2]{Shengdong Du}
\cormark[6]
\author[1,2]{Tianrui Li}

\address[1]{School of Computing and Artificial Intelligence, Southwest Jiaotong University, Chengdu 611756, China} 
\address[2]{Engineering Research Center of Sustainable Urban Intelligent Transportation, Ministry of Education, Chengdu 611756, China} 
\cortext[cor6]{Corresponding author}  

\begin{abstract}
In the era of big data, the issue of data quality has become increasingly prominent. One of the main challenges is the problem of duplicate data, which can arise from repeated entry or the merging of multiple data sources. These "dirty data" problems can significantly limit the effective application of big data. To address the issue of data deduplication, we propose a pre-trained deduplication model based on active learning, which is the first work that utilizes active learning to address the problem of deduplication at the semantic level. The model is built on a pre-trained Transformer and fine-tuned to solve the deduplication problem as a sequence to classification task, which firstly integrate the transformer with active learning into an end-to-end architecture to select the most valuable data for deduplication model training, and also firstly employ the R-Drop method to perform data augmentation on each round of labeled data, which can reduce the cost of manual labeling and improve the model’s performance Experimental results demonstrate that our proposed model outperforms previous state-of-the-art (SOTA) for deduplicated data identification, achieving up to a 28\% improvement in Recall score on benchmark datasets.
\end{abstract}




\maketitle


\section{Introduction}
The amount of digital data in the world is growing explosively. As a result of this “data deluge,” how to manage storage cost-effectively has become one of the most challenging and important tasks in the big data era \cite{1}. Data deduplication can effectively save storage space and significantly improve data quality. Therefore, detecting duplicate data has been a widely discussed and extensively researched topic \cite{2}\cite{3}. The most widely used approach is based on the idea of fuzzy matching to filter data for duplicate values, where a threshold value is defined manually and records with similarity exceeding this threshold value are considered duplicates. This approach relies not only on the choice of attributes used for the fuzzy matching, but also on the choice of thresholds to determine whether the data is duplicated.
\par
\begin{figure}[!htp]
	\begin{center}
		\includegraphics*[width=\linewidth]{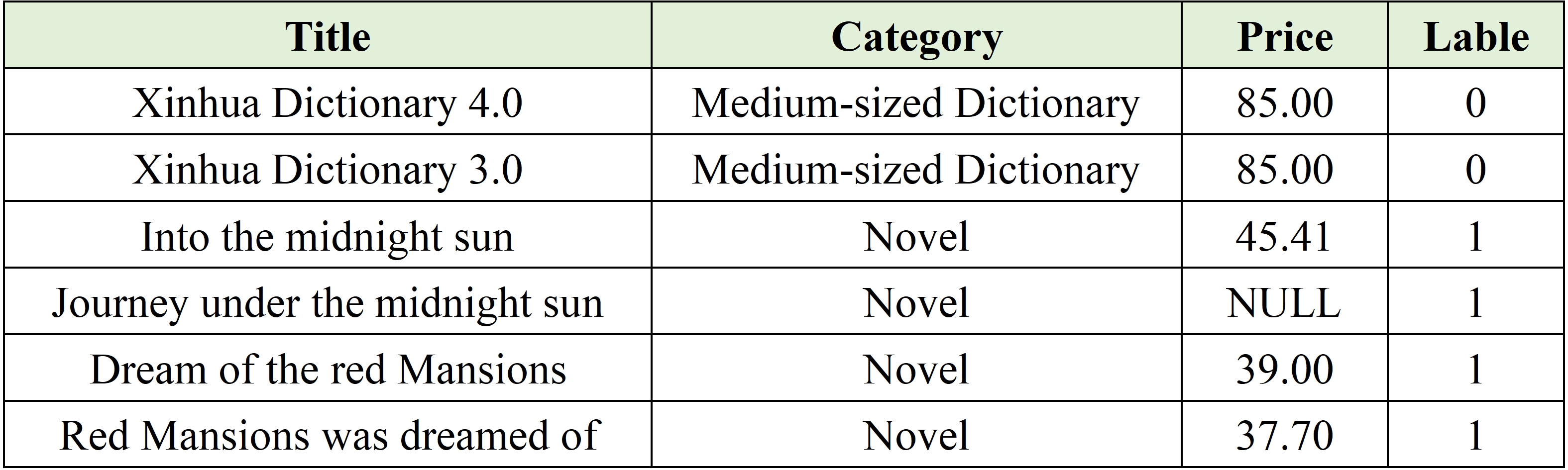}
	\end{center}
	\caption{Examples of duplicates}
	\label{fig0}
\end{figure}

 Duplication occurs for many reasons, such as human manual input errors, duplication during database integration, different sources collect the same object or event in different ways, etc\cite{4}. The existence of these "dirty data" seriously affects the results of data mining, thus affecting decision-making. Figure \ref{fig0} shows an example with duplicate data, where the labels of two adjacent data indicate whether they are repeated, 0 means no repeat, 1 means repeat.

\par As can be seen from the figure \ref{fig0}, in many cases, whether data are duplicates cannot be judged only formally, but must also be understood semantical relation features of different attributes in depth. In the figure, although the first two records are literally similar, they are not duplicates. The reason is that the book editions are different, one is the third edition and the other is the fourth edition. That is, the edition of the book is the important semantic information. The data in the third and fourth rows are duplicates, although they do not appear to be literally similar. For the last two rows of data, although the syntax is different, they are duplicate data, which requires the model to understand the semantic difference. Because of such problems, the traditional similarity or clustering-based duplicate data identification algorithms cannot be fully judged correctly \cite{5}. Therefore, we propose pre-trained data deduplication model based on active learning (PDDM-AL) which can handle the above three situations well. The main reason is that pre-trained models perform well in entity matching and have been shown to understand semantic information well in structured data \cite{5}, and PDDM-AL allows to inject domain knowledge through character marking of more important input fragments, helping the model better understands important information. In addition, R-Drop is also used for data augmentation. which can effectively avoid the semantic changes that classical augmentation method may cause and allows the model to remain stable and more robust even in different situations.

\par We are aware that training a deep learning model typically demands a considerable amount of labeled data due to its intricate model structure and over-parameterization \cite{ren2021survey}. Obtaining an extensive amount of labeled data is often time-consuming \cite{17}; therefore, active learning has emerged as a viable solution to resolve this issue, which is an adaptive machine learning strategy that enables a model to actively and selectively inquire about samples with a high degree of uncertainty or difficulty in classification for labeling, and thereby incorporates these labeled samples into the training set \cite{budd2021survey}. This approach can curtail the number of manually labeled samples while enhancing the generalization capabilities of the model. In contrast to conventional supervised learning, which requires manual labelling of the entire dataset and its utilization to train the model, active learning facilitates the development of training data more efficiently, mitigates overfitting effects, and reduces labeling errors \cite{xie2022towards}. Active learning has widespread applications in various fields such as text processing \cite{schroder2020survey}\cite{zhang2017active}, image categorization \cite{du2019building}\cite{gudovskiy2020deep}\cite{huang2019cost}, and object detection \cite{aghdam2019active}\cite{feng2019deep}\cite{qu2020deep}. In order to reduce the cost of manual labelling, PDDM-AL incorporates active learning techniques based on the use of pre-trained models. The primary concept of this algorithm is to automatically choose the most valuable data from the unlabeled dataset after each training round (selector), particularly when there is a limited amount of labeled data . Subsequently, experts label this selected data and input it back into the model for further training (learner), repeating this process until completion \cite{7}.

\par The main contributions of this paper can be summarized as follows:
\begin{itemize}
	\item We propose a pre-trained deduplication model based on active learning, which is the first work that utilizes active learning to address the problem of deduplication at the semantic level. The model is built on a pre-trained Transformer-based language model and fine-tuned to solve the deduplication problem as a sequence to classification task.
	\item We integrate the transformer with active learning into an end-to-end architecture to select the most valuable data for model training, and also employ the R-Drop method to perform data augmentation on each round of labeled data, which can reduce the cost of manual labeling and improve the model’s performance. 
	\item Experimental results demonstrate that the proposed model outperforms previous stateof-the-art (SOTA) for deduplicated data identification, achieving up to a 28\% improvement in Recall score on benchmark datasets.
\end{itemize}

\par The rest of the paper is organized as follows. Section II surveys the related works. Section III details the proposed PDDM-AL. Section IV reports the experimental results on different datasets and methods. And lastly, Section V summarizes this work.

\section{Related Work}
\label{sec:related}
\par In this section, we briefly introduced some related work, including active learning, data deduplication, and pre-trained models.

\subsection{Active Learning}

\par In real life, the volume of data is usually large and labeling the entire data is undoubtedly time-consuming. The goal of active learning is to find a subset of the unlabelled dataset and manually label them so that the model achieves maximum performance within a fixed labeling budget, rather than labeling the entire data \cite{parvaneh2022active}. Typically, active learning is an iterative process where a portion of data are selected to be labeled from the unlabeled pool and a model is repeatedly trained on the accumulated labeled pool \cite{10}. 

\begin{figure}[!h]
	\begin{center}
		\includegraphics*[width=\linewidth]{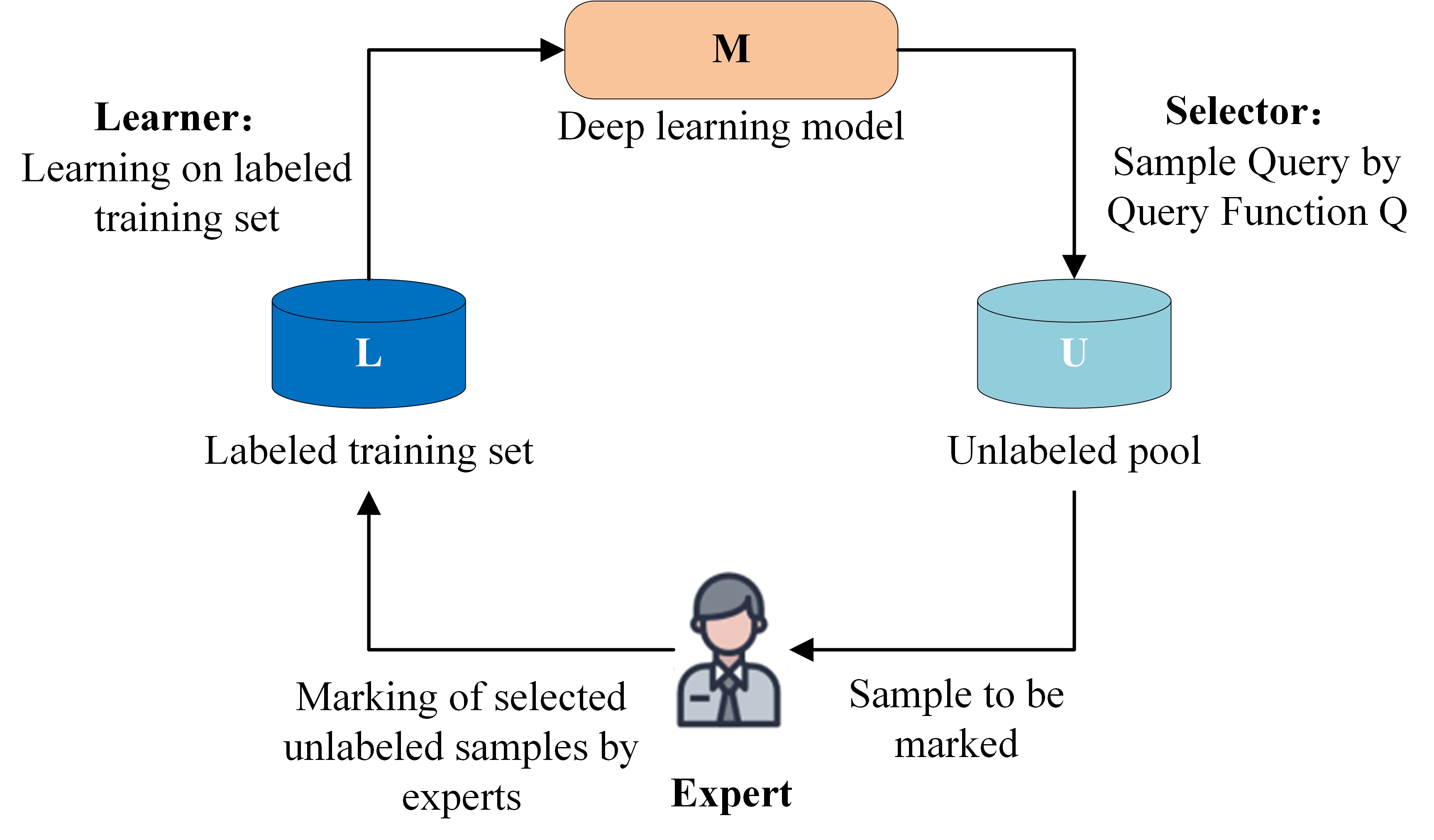}
	\end{center}
	\caption{Standard active Learning Cycle}
	\label{fig10}
\end{figure}

\par As shown in the figure \ref{fig10}, in active learning, its components include the learner and the selector. The learner is responsible for learning the selected data, while the selector is responsible for selecting the most valuable data from the unlabeled data set \cite{20}. For selectors, their main selection strategies include random sampling, uncertain sampling (the most uncertain data), entropy (data with relatively high entropy), and so on. Among them, random sampling is to randomly select samples from the Unlabeled pool for experts to label, which is the simplest method \cite{qayyumi2022active}; Uncertainty sampling is to select data records based on the uncertainty of the model's discrimination of the data record. Generally speaking, data records with higher uncertainty are handed over to manual labeling, that is, the higher the degree of uncertainty, the greater the value contained in the data \cite{russo2020active}. Entropy-based sampling selects sample data with relatively high entropy to be labeled by humans \cite{wu2022entropy}. The most widely-used AL strategy is uncertainty sampling due to its surprisingly simple but very effective performance.

\subsection{Data Deduplication}

\par Among the deduplication algorithms, the commonly used methods include fuzzy matching algorithms, SNM-based algorithms \cite{11}, clustering-based algorithms, and neural network-based algorithms. 
\par The algorithm based on fuzzy matching follows the principle of selecting one or more keywords and calculating the similarity between each data and other data using a string similarity formula. If the similarity exceeds a threshold, the data is considered a duplicate. Commonly used algorithms for this purpose include the Smith-Waterman algorithm and the N-Gram algorithm \cite{12}. The SNM-based algorithm is mainly divided into three steps: selecting keywords, sorting, and merging \cite{11}. But there is a problem with the above two methods: it is difficult to select keywords. If the keyword selection is not appropriate, it will not only directly affect the detection efficiency, but also have a great impact on the accuracy of the results. The main idea of clustering-based methods, such as Dedupe \cite{13}, is to first use support vector machines to achieve accurate distance estimation between records composed of multiple fields, and finally through hierarchical clustering, the distance metric adopts the idea of "average metric", and the Groups that are close to each other are grouped together \cite{13}. Neural network-based algorithms such as Sentencebert: a variant of the BERT model that trains sentence embeddings input for sentence similarity search, the trained model generates a high dimensionality for each record (e.g. 768 for BERT) vector, use its cosine similarity to the other vectors to find duplicates \cite{14}.

\subsection{Pretrained Model}
\par Transformer-based Pretrained Language Models (TPLMs), such as BERT \cite{8} and RoBERTa \cite{9}, have demonstrated good performance on a wide range of NLP tasks, and Pre-training with a large corpus of unlabeled text, such as Wikipedia \cite{19}. It has been shown that after pre-training, the shallow layer captures the syntactic and semantic meaning of the input sequence \cite{clark2019does}\cite{tenney2019bert}. 
\par Considering that the transformer architecture calculates token embeddings using all tokens in the input sequence, the embeddings generated by pre-trained models are more contextually relevant compared to other methods like word2vec, GloVe, or FastText. Furthermore, they have a deeper understanding of the semantics \cite{18}. Therefore, embeddings obtained from pre-trained models can capture polysemous words, i.e., identify that the same word may have different meanings in different phrases; such models can also understand the opposite case, where different words may have the same meaning. This powerful language understanding ability can greatly improve the performance of data deduplication. In entity matching, pre-training models have been shown to understand contextual information well and perform well in dirty datasets \cite{5}\cite{7}.

\section{Methodology}

\begin{figure*}[!htp]
	\begin{center}
		\includegraphics*[width=\textwidth]{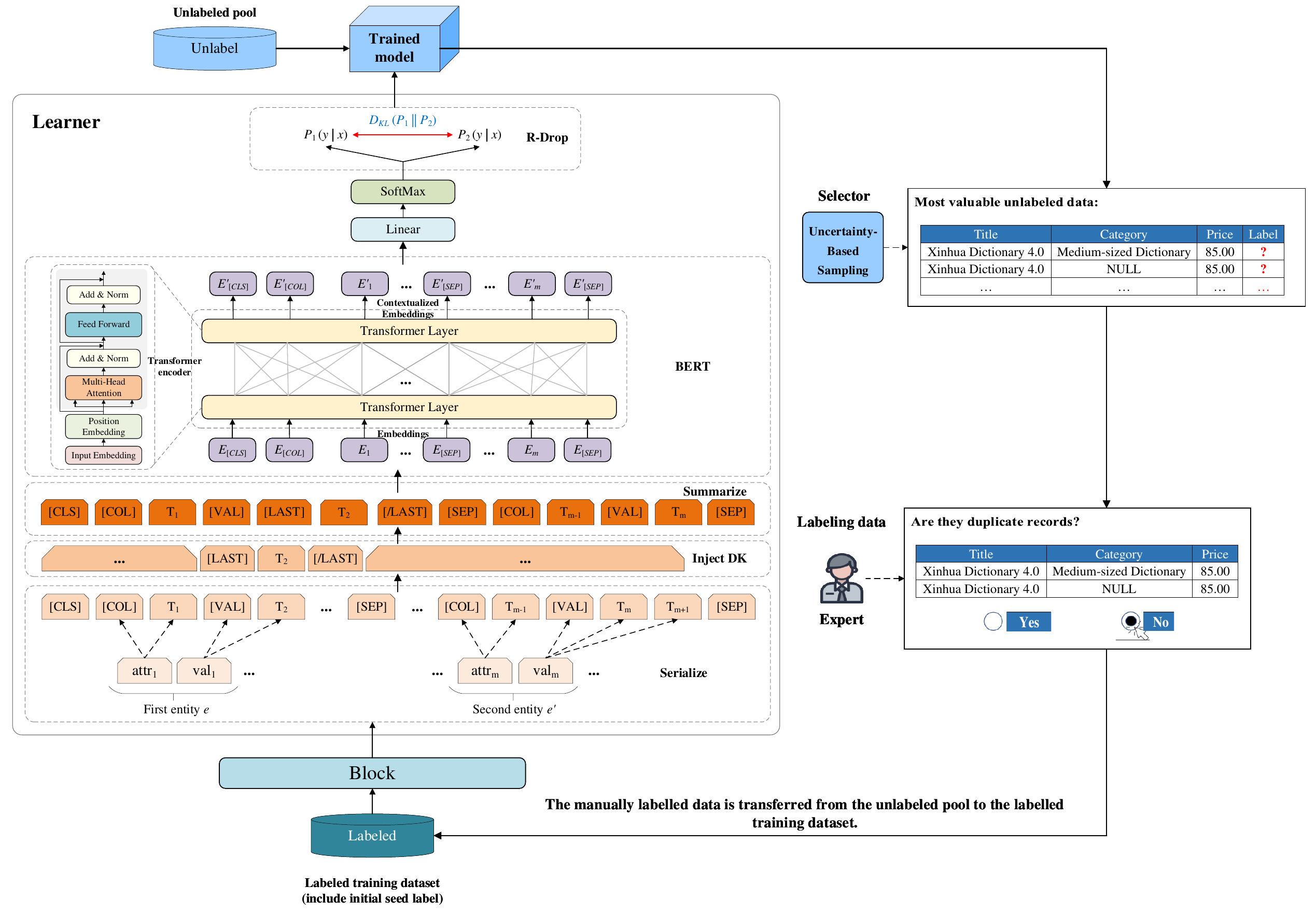}
	\end{center}
	\caption{The design diagram of PDDM-AL}
	\label{fig1}
\end{figure*}

The overall architecture of PDDM-AL, which is shown in Figure \ref{fig1}. The model firstly introduce active learning as the basic framework, and we fine-tune the classification problem by treating deduplication as a sequence for the first time. The main process of model training is as follows: First, the data is blocked to exclude pairs that are unlikely to be duplicate records (e.g. if no word is the same between two records, they are considered unlikely to be duplicates), and candidate pairs are generated by the block strategy. For candidate pairs, serialized, injection of domain knowledge and summary operations are performed in turn. The above processed data is then fed into the model for training (while the data is augmented by R-Drop). Once training is complete, the selector feeds the unlabelled data into the trained model for prediction, obtaining predictions of 0 and 1 (1 for duplicate data and 0 for non-duplicate data), while the selector will select out the data that the model is most uncertain for expert labelling. Finally, the expert-labelled data is transferred to the labelled dataset to continue the active learning iteration. Details of the above computing process are described below.

\subsection{Preprocessing}
\par PDDM-AL first uses the Block strategy to generate candidate pairs (the candidate pair contains two records that may be duplicates). The block strategy used in this paper is that at least one word is the same between candidate pairs. For candidate pairs ($e$,$e\prime$), they need to be serialised into data that can be used by the model for classification. Each entity ${\rm{e}} = {\{ att{r_i},va{l_i}\} _{0 < i \le m}}$ is serialised according to the following formula, i.e. [COL] is prefixed to the attribute name and [VAL] is prefixed to the attribute value.

\begin{equation}
	\begin{split}
		{\rm{serialize}}(e) = [{\rm{COL}}]att{r_1}[{\rm{VAL}}]va{l_1}\\...[{\rm{COL}}]att{r_{\rm{m}}}[{\rm{VAL}}]va{l_m},
		\label{pythagorean}
	\end{split}
\end{equation}
Where $m$ represents the number of attributes of e. For example, the first piece of data in Figure \ref{fig0} is serialized as: "[COL] title [VAL] Xinhua Dictionary 4.0 [COL] category [VAL] Medium-sized dictionary [COL] price [VAL] 85.00".

For candidate pairs ($e$,$e\prime$), we let

\begin{equation}
	\begin{split}
		{\rm{serialize}}(e,e\prime ) = {\rm{[CLS]serialize}}(e){\rm{[SEP]}}\\{\rm{serialize}}(e\prime ){\rm{[SEP]}}
		\label{pythagorean}
	\end{split}
\end{equation}
Where the [CLS] is a special token for BERT, and this vector will be fed into the fully connected layer for classification.

\par The next step is to inject domain knowledge into the serialised data to highlight which pieces of information may be important. As described in the first section, version information is particularly important for books, even though it is literally just a number. PDDM-AL uses an open-source Named Entity Recognition (NER) model \cite{16} to identify known types such as person, date or organization and uses regular expressions to identify specific types such as version Information, a phone number digits, etc. For the serialized serialize(e, e'), input it into the NER model and get the following output:

\begin{equation}
	\begin{split}
		\{ start,end,label\}  = {\rm{NER}}({\rm{serialize}}({\rm{e}},{\rm{e'}}))
		\label{e10}
	\end{split}
\end{equation}
Where the NER model uses the en\_core\_web\_lg model in spacy \cite{16}. $start$ represents the starting position of the entity, $end$ represents the end position of the entity, $label$ is the label of the recognized entity. For each entity identified by NER and a regular expression, use [LAST] [/LAST] to mark the identified key points, as shown in Equation \ref{e3} and Equation \ref{e4}.

\begin{equation}
	\begin{split}
		{\rm{serialize}}{({\rm{e}},{\rm{e'}})_{{\rm{start}}}} = [{\rm{LAST}}],
		\label{e3}
	\end{split}
\end{equation}

\begin{equation}
	\begin{split}
		{\rm{serialize}}{({\rm{e}},{\rm{e'}})_{{\rm{end}}}} = [{\rm{/LAST}}],
		\label{e4}
	\end{split}
\end{equation}
Where $start$ is the starting position of the entity and $end$ is the ending position of the entity.
For example after the above process, the version information of the first piece of data in Figure \ref{fig0} is identified and marked, it will become "..[COL] title [VAL] Xinhua Dictionary [LAST] 4.0 [/LAST].."
\par The final step in data processing is to summarise, as it is not conducive to the model learning important knowledge if the serialised entries are too long, and the Bert model has its own input limit (up to 512 subword tokens). PDDM-AL chooses to keep non-stop word tokens with higher TF-IDF scores, as shown in Equation \ref{e5}.

\begin{equation}
	\begin{aligned}
		{x}{\rm{ = MAX\_TFIDF}}{({\rm{serialize}}(e,e'))_j},{\rm{0}} < j < n,
		\label{e5}
	\end{aligned}
\end{equation}
Where $x$ represents the serialized entries obtained after preprocessing, as the input of Bert model. $n$ represents the maximum string length of the model input.

\subsection{Pre-trained Model with Active Learning}
\par The fourth step involves feeding the pre-processed string into the pre-trained model BERT, which utilizes the Transformer architecture to calculate token embeddings from all tokens in the input sequence. As a result, the generated embeddings are contextually rich and effectively capture the semantic and contextual nuances of the words. The formula used for attention calculation is shown below.

\begin{equation}
	\begin{aligned}
		{\rm{Attention(Q,K,V) = softmax(}}\frac{{Q{K^T}}}{{\sqrt {{d_k}} }}{\rm{)V}}
		\label{e11}
	\end{aligned}
\end{equation}

Where Q represents the query vector corresponding to each element(Query), K represents the key vector corresponding to each element(Key), and V represents the value vector corresponding to each element(Value), all of which are the results of different linear transformations of the original sequence $x$ and can be used as representatives of $x$. The implication is a weighted linear combination of word vectors that have been expressed mathematically using attention weights, so that each word vector contains information about all the word vectors within the current sentence. 

In the Transformer model, the self-attention layer is enhanced through a multi-headed attention mechanism. The process involves mapping Query, Key, and Value using h different linear transformations. Subsequently, the individual attentions are combined, followed by another linear transformation. The entire calculation process can be represented as follows:

\begin{equation}
	\begin{aligned}
		{\rm{MultiHead}}(Q,K,V) = Concat(hea{d_1},...,hea{d_h}){W^0}
		\label{e11}
	\end{aligned}
\end{equation}

\begin{equation}
	\begin{aligned}
		hea{d_i} = {\rm{Attention}}(QW_i^Q,KW_i^K,VW_i^V)
		\label{e11}
	\end{aligned}
\end{equation}

\par Through the calculation of the above process, each set of attention is used to map the input to a different sub-representation space, which allows the model to focus on different locations in different sub-representation spaces.

To make the pre-trained model suitable for binary classification tasks, we add a simple fully connected layer and a softmax output layer after the last layer of Bert, and use the [CLS] output of the Bert model as the input of the fully connected layer, The formula is as follows. 

\begin{equation}
	\begin{aligned}
		r = {\rm{Bert\_CLS}}(x)
		\label{e11}
	\end{aligned}
\end{equation}

\begin{equation}
	\begin{aligned}
		{\rm{p}} = {\rm{softmax}}(wr + b)
		\label{e12}
	\end{aligned}
\end{equation}

Where $x$ is the data obtained after preprocessing, $r$ stands for [CLS] of Bert output, $w$ and $b$ are the weights and biases of the linear layer.

The pre-trained Bert model is fine-tuned with the above labeled and preprocessed data, and the R-Drop\cite{15} strategy is used for data augmentation which can help models perform better on dirty data, such as null values, misspellings, etc. However, as described above, traditional data augmentation methods, such as changing or deleting words, can easily cause semantic changes and cause the model to learn wrong information. Since R-Drop\cite {15} discards some neurons at random each time, making it possible to change the output vector while minimizing semantic changes.
\par  For the training data $D$=\{$x_i$,$y_i$\} each training sample $x_i$ will be forward propagated through the network twice. Then we get two output predictions: $p_1$($y_i$\textbar $x_i$) and $p_2$($y_i$\textbar $x_i$) , because Dropout randomly drops a part of neurons each time, so $p_1$($y_i$\textbar $x_i$) and $p_2$($y_i$\textbar $x_i$) are two different predicted probabilities obtained by two different subnetworks (from the same model). R-Drop employs a symmetric Kullback-Leibler (KL) divergence to constrain $p_1$($y_i$\textbar $x_i$) and $p_2$($y_i$\textbar $x_i$) by applying these two different predicted probabilities:

\begin{equation}
	\begin{split}
		{L_{KL}^i} = \frac{1}{2}({D_{KL}}({P_1}({y_i}|{x_i})||{P_2}({y_i}|{x_i})) \\+ {D_{KL}}({P_2}({y_i}|{x_i})||{P_1}({y_i}|{x_i}))),
		\label{pythagorean}
	\end{split}
\end{equation}

\par Then the classic maximum likelihood loss function is expressed as follows:

\begin{equation}
	\begin{split}
		L_{NLL}^i =  - \log {P_1}({y_i}|{x_i}) - \log {P_2}({y_i}|{x_i}),
		\label{pythagorean}
	\end{split}
\end{equation}
\par The final training loss function is:

\begin{equation}
	\begin{split}
		{L_i} = L_{NLL}^i + \alpha  * L_{KL}^i,
		\label{pythagorean}
	\end{split}
\end{equation}

where $\alpha$ is the coefficient used to control $L_{KL}^i$.
\par After the training described above, the selector will select the data that the current model is most uncertain about (the closer the model's output probability p is to 0.5, the more uncertain the model is), and send it to the expert for labeling. The expert-labeled data will enter the labeled data sets and continue to iterate by repeating the above steps. The main active learning process is as follows:

\begin{algorithm}[h]
	\renewcommand{\algorithmicrequire}{\textbf{Input:}}
	\renewcommand{\algorithmicensure}{\textbf{Output:}}
	\caption{Active Learning Algorithms} 
	\begin{algorithmic}[1]
		\Require 
		Labeled dataset T={($e$,$e\prime$,label)}, unlabeled dataset R, the number of active learning cycles N
		\Ensure The trained model M, data after model de-duplication
		\State $i \gets 0$.
		\While{i \textless N} 
		\State Serialize the data T according to the equations (1) and (2).
		\State Inject domain knowledge into the serialised data according to equations (3), (4) and (5).
		\State The data were summarized according to equation (6) to obtain the pre-processed data $x$.
		\State Put the above pre-processed data $x$ into Model for fine-tuning and R-drop according to equations (12), (13), (14).
		\State $r$ $\gets$ ${\rm{Bert\_CLS}}(x)$.
		\State $confidence$ $\gets$ ${\rm{softmax}}(wr + b)$.
		\State $n\_samples$ $\gets$ least $confidence$ n samples in R.
		\State $r\_label$ $\gets$ experts mark $n\_samples$ .
		\State $T$ $\gets$ $r\_label$.
		\State $i\gets i+1$.
		\EndWhile
	\end{algorithmic}
\end{algorithm}

\section{Experiments}
\par In this chapter, we validate the proposed model on several real data sets. The experiments in this chapter are divided into two main parts.
\par The first part involves comparing the Precision, Reacll, and F1 of the PDDM-AL model with other data de-duplication models to verify its data de-duplication abilities.
The second part of the experiments involves two main aspects. Firstly, we compare the F1 and Recall of the active learning strategy for each round. Secondly, we conduct experiments and comparisons between the uncertainty-based sampling strategy and random sampling to validate the correctness of the active learning strategy utilized by the model.

\subsection{Datasets and baseline model}
We have experimented on four real data sets. The algorithm of this type of model is designed for entities in multi-source data, with the purpose of grouping all the matching entities from different sources into entity clusters. The following dataset is used in this experiment:
\par 1. The Music Brainz dataset is based on real records about songs from the MusicBrainz database but uses the DAPO data generator to create duplicates with modified attribute values. The generated dataset consists of five sources and contains duplicates for 50\% of the original records in two to five sources. All duplicates are generated with a high degree of corruption to stress-test the model. 
\par 2. Geographic Settlements contains geographical real-world entities from four different data sources (DBpedia, Geonames, Freebase, NYTimes) and has already been used in the OAEI competition.
\par 3. The Education dataset was obtained from a list of 10 different sources. 
\par 4. The enterprise personnel dataset is provided by the Federation of Industry and Commerce, which is Chinese data.
\par The training, testing, and validation sets of the four datasets mentioned above are partitioned into a ratio of 6:2:2. The details of the data are shown in the table \ref{tbl1}.

\begin{table*}[width=1\textwidth,pos=!h]
	\caption{Description of the experimental dataset}\label{tbl1}
	\begin{tabular*}{\tblwidth}{@{} LLLL@{} }
		\toprule
		Dataset name & Number of attributes & Sample size & Field\\
		\midrule
		Musicbrainz-20-A01$^{\mathrm{a}}$               & 12                   & 19375       & Music \\
		GeographicSettlements$^{\mathrm{a}}$            & 9                    & 3054        & Ggeography \\
		Education                        & 33                   & 3337        & Educate \\
		Enterprise personnel information & 19                   & 6002        & Personnel information \\
		\bottomrule
	\end{tabular*}
\par{$^{\mathrm{a}}$Source: The data comes from the public dataset website Benchmark datasets for entity resolution: https://dbs.uni-leipzig.de/de.}
\end{table*}

\subsection{Baseline model and evaluation indicators}
In order to verify the data deduplication performance of the model, the PDDM-AL proposed in this paper is compared with three benchmark algorithms: Sentencebert \cite{14}, Dedupe \cite{13} and field-based similarity. 
\par Sentencebert \cite{14}: One of the latest techniques for detecting duplicate data is a bert-based model variant that trains sentence embeddings for comparing sentence similarity. The trained model produces a high-dimensional vector for each record, and determines duplicates by measuring cosine similarity between vectors.
\par Dedupe \cite{13}: One of the advanced techniques for identifying duplicate data involves using svm to accurately estimate the distance between records consisting of multiple fields. finally, through hierarchical clustering, the distance metric adopts the concept of "average metric" to group together those groups that are close to each other.
\par Field-based similarity: One of the traditional methods of identifying duplicate data, calculating the similarity of specified fields and identifying data with high similarity as duplicate data.

\par The ratio of training set, validation set, and test set is 6:2:2. We use 12 layers Bert, in all experiments, $\alpha$ takes the value 0.8, and the batch size is 34.
\par The criteria used to judge the performance of the model are Recall, Precision and F1 value, which are calculated as follows:
\begin{equation}
	\begin{split}
		Recall = \frac{{TP}}{{TP + FN}}
		\label{pythagorean}
	\end{split}
\end{equation}

\begin{equation}
	\begin{split}
		Precision = \frac{{TP}}{{TP + FP}}
		\label{pythagorean}
	\end{split}
\end{equation}

\begin{equation}
	\begin{split}
		F1 = \frac{{2Precision*Recall}}{{Precision + Recall}}
		\label{pythagorean}
	\end{split}
\end{equation}
where $TP$ denotes true positive, which refers to the number of records that indicate the same entity and are deemed accurate. $FN$ represents false negative, which corresponds to the number of non-duplicated records that are judged incorrect. $FP$ stands for false positive, meaning the number of records that do in fact denote the same entity as other records, but are incorrectly deemed otherwise.

\subsection{Analysis of experimental results}
The experiment is mainly divided into two parts. The first part uses the same amount of labeled data to train on each model to verify the effectiveness of PDDM-AL in data deduplication. In the second part of the experiment, a small amount of labelled data is fed into PDDM-AL for training and then enters the active learning loop, adding small amounts of data in turn until convergence, verifying that active learning can rapidly improve the accuracy of the model with a small number of labels.
\par First of all, for the first part of the experiment, we use all the data in the training set to train the model until convergence, the results obtained are reported in Table \ref{tbl2}. PDDM-AL generally outperforms other methods in terms of Precision, Recall and F1 scores. However, in the education dataset, its precision was 0.0483 lower than Dedupe, resulting in a slight lag in F1 score by 0.0065. Nonetheless, PDDM-AL still achieved a higher recall rate by 0.0322 than Dedupe. Compared to Sentencebert, PDDM-AL's F1 score is consistently higher, with an increase of 0.009, 0.294, 0.246, and 0.056 in respective datasets. Additionally, PDDM-AL also shows improvement in recall rates, which are increased by 0.023, 0.287, 0.267, and 0.192, respectively. Although PDDM-AL's precision is slightly lower than Sentencebert's in the Musicbrainz-20-A01 dataset by 0.02, it performs better in the other three datasets with higher precision scores of 0.3013, 0.2237, and 0.1569, respectively.

\begin{table*}[width=1\textwidth,pos=!h]
	\centering
	\caption{Comparison of data deduplication results of different models}
	\label{tbl2}
	\resizebox{\linewidth}{!}{
		\begin{tblr}{
				cells = {c},
				cell{1}{1} = {r=2}{},
				cell{1}{2} = {c=3}{},
				cell{1}{5} = {c=3}{},
				cell{1}{8} = {c=3}{},
				cell{1}{11} = {c=3}{},
				hline{1,3,7} = {-}{},
				hline{2} = {2-13}{},
			}
			Model method           & Musicbrainz-20-A01 &                 &                 & GeographicSettlements &                 &                & Education       &                &                 & {Enterprise personnel \\ information} &                &                 \\
			& Precision          & Recall          & F1              & Precision             & Recall          & F1             & Precision       & Recall         & F1              & Precision                             & Recall         & F1              \\
			Sentencebert           & \textbf{0.9603}    & 0.9694          & 0.9648          & 0.6674                & 0.6664          & 0.6669         & 0.7137          & 0.6742         & 0.6934          & 0.8371                                & 0.8063         & 0.8214          \\
			Dedupe                 & 0.9468             & 0.6594          & 0.7774          & \textbf{0.9894}       & 0.9257          & 0.9565         & \textbf{0.9857} & 0.9088         & \textbf{0.9457} & 0.7820                                & 0.9972         & 0.8766          \\
			Field-based similarity & 0.5737             & 0.6517          & 0.6102          & 0.7661                & 0.8416          & 0.8021         & 0.4976          & 0.7301         & 0.5918          & 0.8339                                & 0.8600         & 0.8468          \\
			PDDM-AL                & 0.9583             & \textbf{0.9917} & \textbf{0.9747} & 0.9687                & \textbf{0.9534} & \textbf{0.9610} & 0.9374          & \textbf{0.9410} & 0.9392          & \textbf{0.9940}                        & \textbf{0.9980} & \textbf{0.9979} 
	\end{tblr}}
\end{table*}

\par
\begin{figure}[!htp]
	\begin{center}
		\includegraphics*[width=1\linewidth]{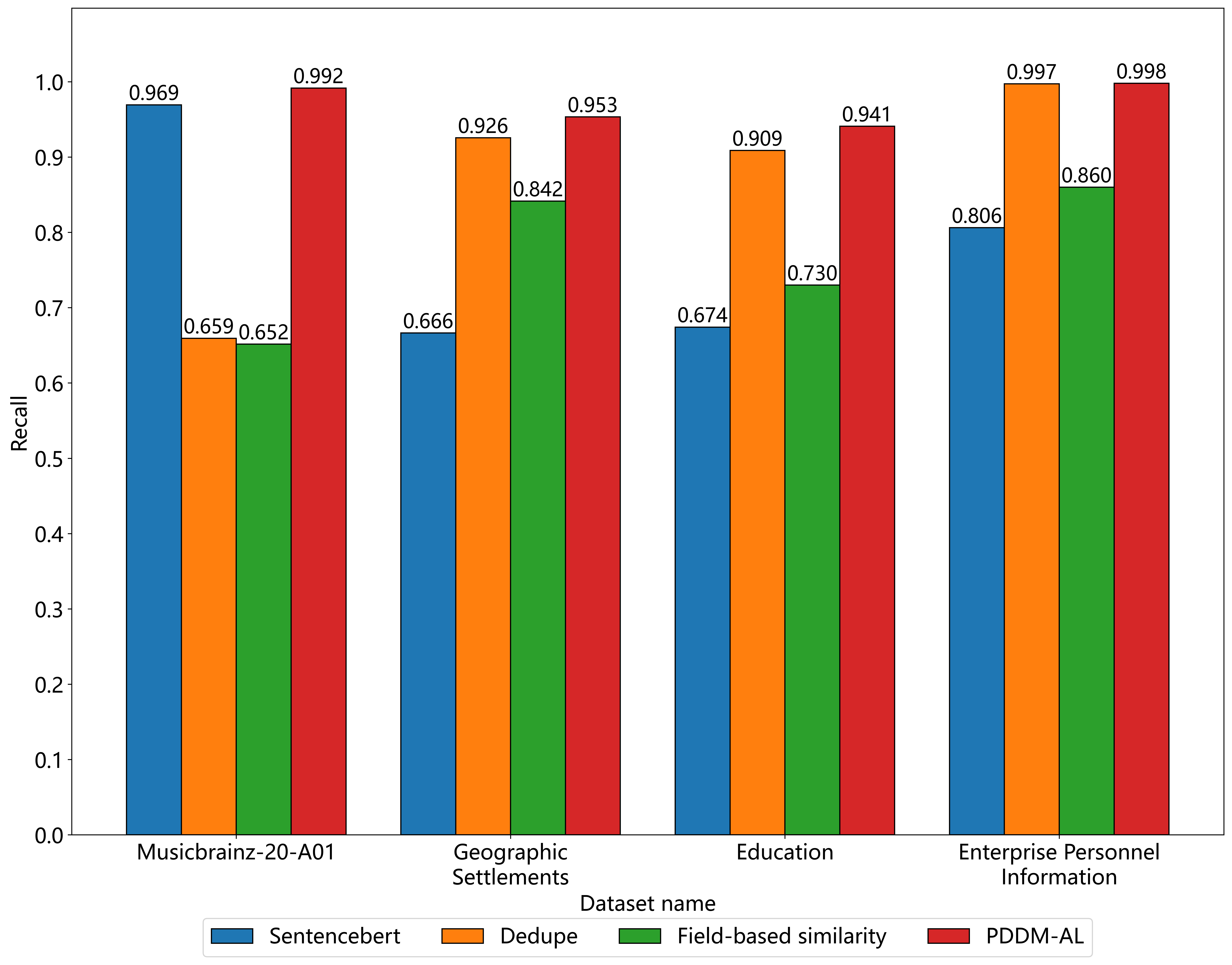}
	\end{center}
	\caption{Histogram of deduplication Recall of different models}
	\label{fig2}
\end{figure}

\begin{figure}[!htp]
	\begin{center}
		\includegraphics*[width=1\linewidth]{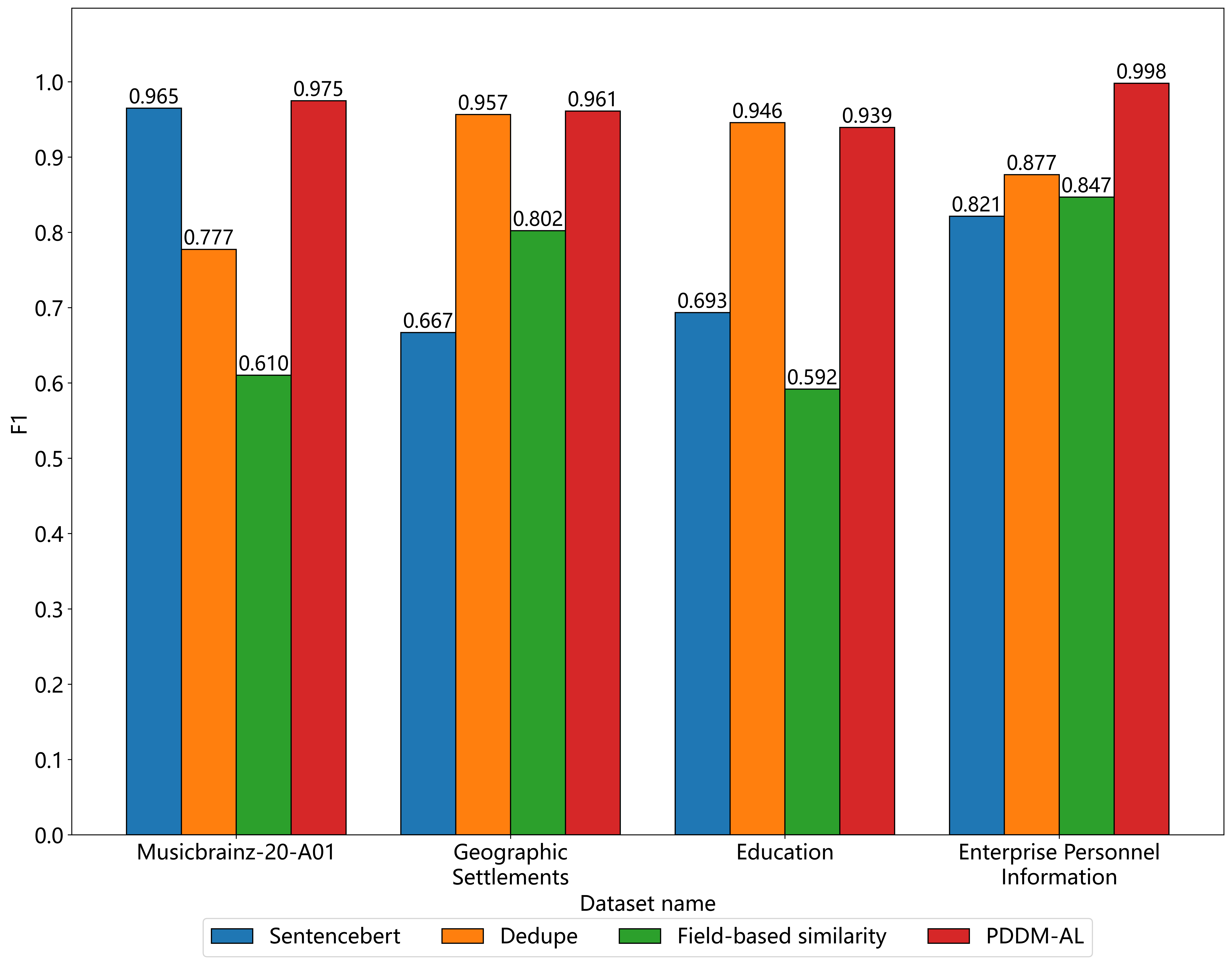}
	\end{center}
	\caption{Histogram of deduplication F1 of different models}
	\label{fig7}
\end{figure}

 According to figure \ref{fig2} and figure \ref{fig7}, it can be seen that on the four data sets, the traditional method Field-based similarity does not work very well. This is mainly because this method is very dependent on the selection of key attributes and its data quality. If the key attributes are not selected properly or if there is a high presence of dirty data, the performance of the model can be significantly compromised. At the same time, it can be seen from the figure that Sentencebert has poor performance on the GeographicSettlements and Education datasets, mainly because these two datasets contain more missing values and useless attributes, and Sentencebert is not very good at extracting key information in the dataset. Compared with the baseline models, PDDM-AL has achieved certain improvements in four data sets.

\par For the second part of the experiment, the main purpose is to verify the effect of active learning, so the F1 and Recall of the model after each round of active learning are recorded in turn (one round of active learning will select 1000 pairs of data for labeling and then input to the model training). The main results are as follows.

\begin{table*}[width=1\textwidth,pos=!h]
	\caption{F1 of the three datasets at each round of active learning}\label{tbl3}
	\begin{tabular*}{\tblwidth}{@{} LLLLLL@{} }
		\toprule
		Dataset               & First round & Second round & Third round & Fourth round & Fifth round\\
		\midrule
		Musicbrainz-20-A01    & 0.674       & 0.948        & 0.973       & 0.973        & 0.974        \\
		GeographicSettlements & 0.798       & 0.936        & 0.961       & 0.961        & 0.961        \\
		Education             & 0.846       & 0.939        & 0.887       & 0.887        & 0.887        \\
		\bottomrule
	\end{tabular*}
\end{table*}

\begin{table*}[width=1\textwidth,pos=!h]
	\caption{Recall of the three datasets at each round of active learning}\label{tbl4}
	\begin{tabular*}{\tblwidth}{@{} LLLLLL@{} }
		\toprule
		Dataset               & First round & Second round & Third round & Fourth round & Fifth round\\
		\midrule
		Musicbrainz-20-A01    & 0.970       & 0.924        & 0.989       & 0.989        & 0.992        \\
		GeographicSettlements & 0.708       & 0.935        & 0.953       & 0.953        & 0.953        \\
		Education             & 0.891       & 0.941        & 0.900       & 0.900        & 0.900        \\
		\bottomrule
	\end{tabular*}
\end{table*}

As can be seen from Table \ref{tbl3} and Table \ref{tbl4}, active learning enables the model to be trained with a small amount of labeled data, which can quickly improve the F1 and Recall. For example, on the GeographicSettlements dataset, between the first and second rounds, only 1,000 pairs of data were added for training, so that the F1 and Recall of the model increased by 0.138 and 0.227. On three datasets, the model outperforms other models using only 3000 pairs of data. This is mainly because the active learning strategy in PDDM-AL can effectively select the most valuable data for the model and hand it over to experts for labeling through the selector, so that allows training with a small amount of the most valuable data and rapidly improves the model's performance.

\begin{table*}[htbp]
	\centering
	\caption{F1 for different active learning selection strategies on three datasets}
	\label{tbl6}
	\resizebox{\linewidth}{!}{
		\begin{tblr}{cells = {c},
				cell{1}{1} = {r=2}{},
				cell{1}{2} = {c=2}{},
				cell{1}{4} = {c=2}{},
				cell{1}{6} = {c=2}{},
				hline{1,3,8} = {-}{},
				hline{1,3,8} = {-}{},
				hline{2} = {2-7}{},
			}
			Number of data pairs & Musicbrainz-20-A01 &                  & GeographicSettlements &                  & Education   &                  \\
			& Uncertainty        & Random selection & Uncertainty           & Random selection & Uncertainty & Random selection \\
			1000         & 0.674              & 0.674            & 0.798                 & 0.798            & 0.846       & 0.846            \\
			2000         & 0.948              & 0.891            & 0.936                 & 0.932            & 0.939       & 0.890            \\
			3000         & 0.973              & 0.910            & 0.961                 & 0.931            & 0.887       & 0.834            \\
			4000         & 0.973              & 0.910            & 0.961                 & 0.931            & 0.887       & 0.834            \\
			5000         & 0.974              & 0.910            & 0.961                 & 0.931            & 0.887       & 0.834            
	\end{tblr}}
\end{table*}

\begin{table*}[htbp]
	\centering
	\caption{Recall for different active learning selection strategies on three datasets}
	\label{tbl7}
	\resizebox{\linewidth}{!}{
		\begin{tblr}{
				cells = {c},
				cell{1}{1} = {r=2}{},
				cell{1}{2} = {c=2}{},
				cell{1}{4} = {c=2}{},
				cell{1}{6} = {c=2}{},
				hline{1,3,8} = {-}{},
				hline{2} = {2-7}{},
			}
			Number of data pairs & Musicbrainz-20-A01 &                  & GeographicSettlements &                  & Education   &                  \\
			& Uncertainty        & Random selection & Uncertainty           & Random selection & Uncertainty & Random selection \\
			1000                 & 0.970              & 0.970            & 0.708                 & 0.708            & 0.891       & 0.891            \\
			2000                 & 0.924              & 0.989            & 0.935                 & 0.929            & 0.941       & 0.905            \\
			3000                 & 0.989              & 0.980            & 0.953                 & 0.942            & 0.900       & 0.856            \\
			4000                 & 0.989              & 0.980            & 0.953                 & 0.942            & 0.900       & 0.856            \\
			5000                 & 0.992              & 0.980            & 0.953                 & 0.942            & 0.900       & 0.856            
	\end{tblr}}
\end{table*}

\begin{figure}[!htp]
	\begin{center}
		\includegraphics*[width=1\linewidth]{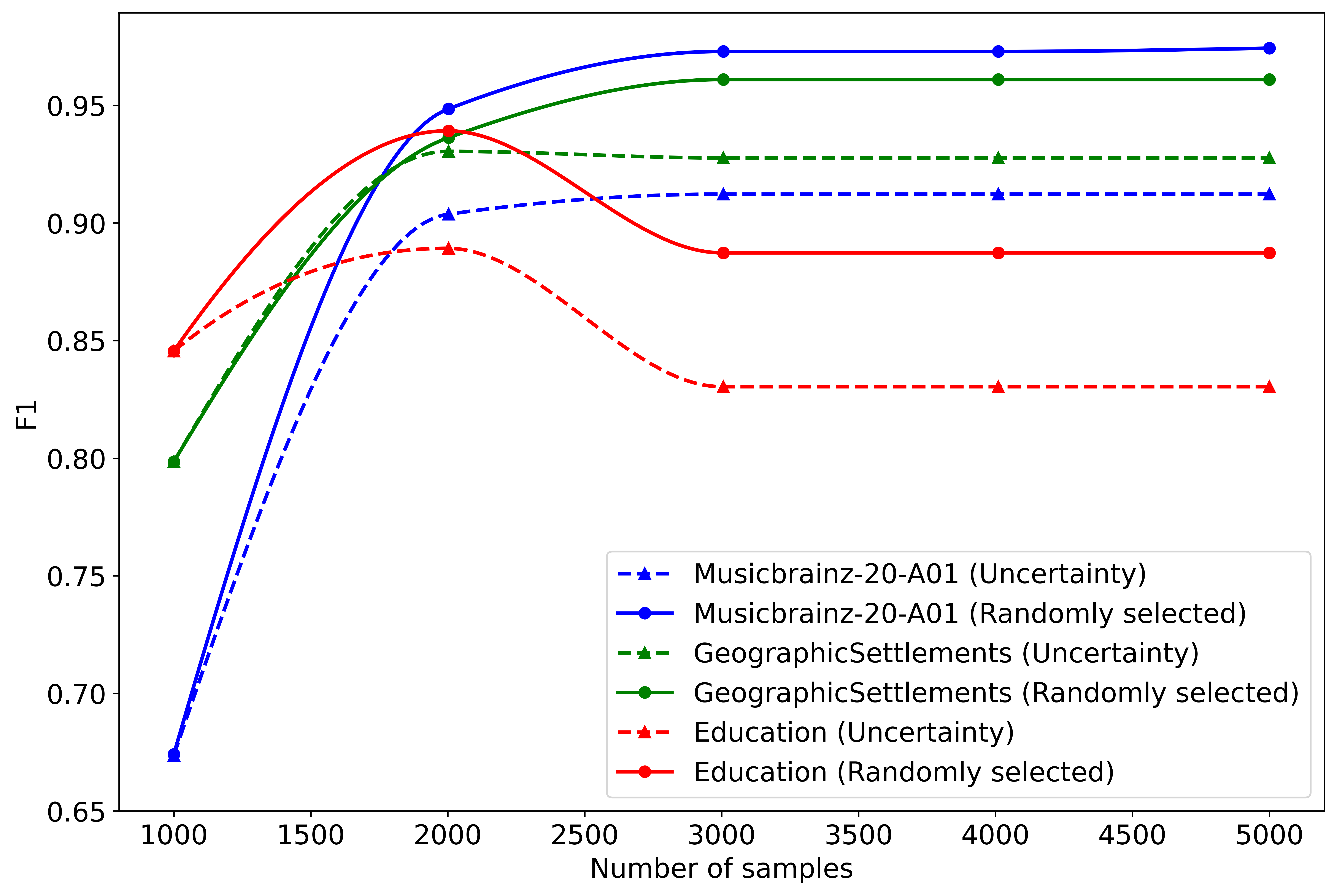}
	\end{center}
	\caption{Line chart of F1 comparison based on uncertainty strategy and random selection strategy}
	\label{fig3}
\end{figure}

\begin{figure}[!htp]
	\begin{center}
		\includegraphics*[width=1\linewidth]{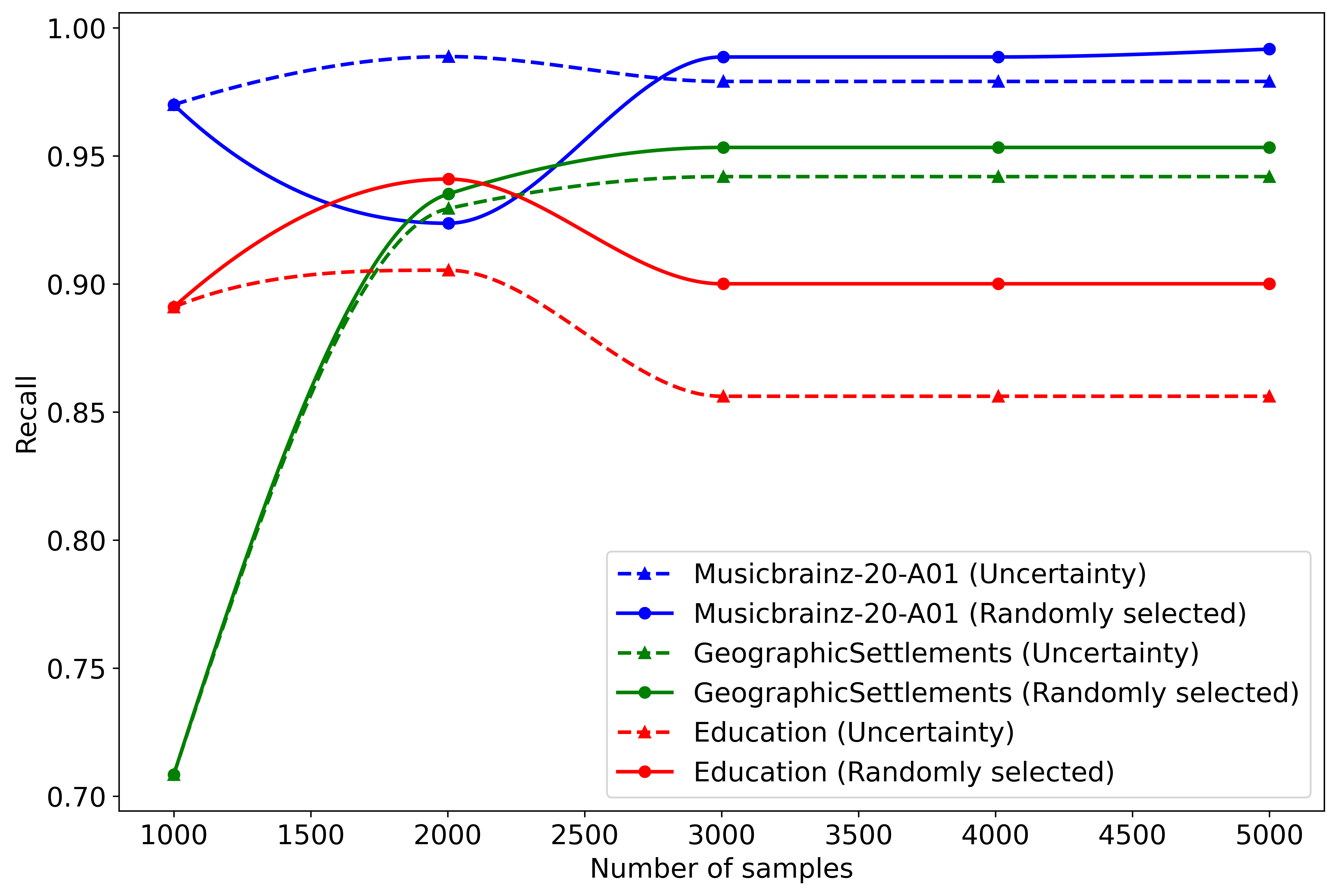}
	\end{center}
	\caption{Line chart of Recall comparison based on uncertainty strategy and random selection strategy}
	\label{fig4}
\end{figure}

From Figure \ref{fig3} and Figure \ref{fig4}, we can see that the selection strategy based on uncertainty is obviously better than random selection. Higher F1 values and recall rates can be obtained with the same amount of data, which proves that the uncertainty selection strategy can select the most valuable data in the current model and improve the F1 and recall rates of the model.

\par
\begin{figure}
	\begin{center}
		\includegraphics*[width=1\linewidth]{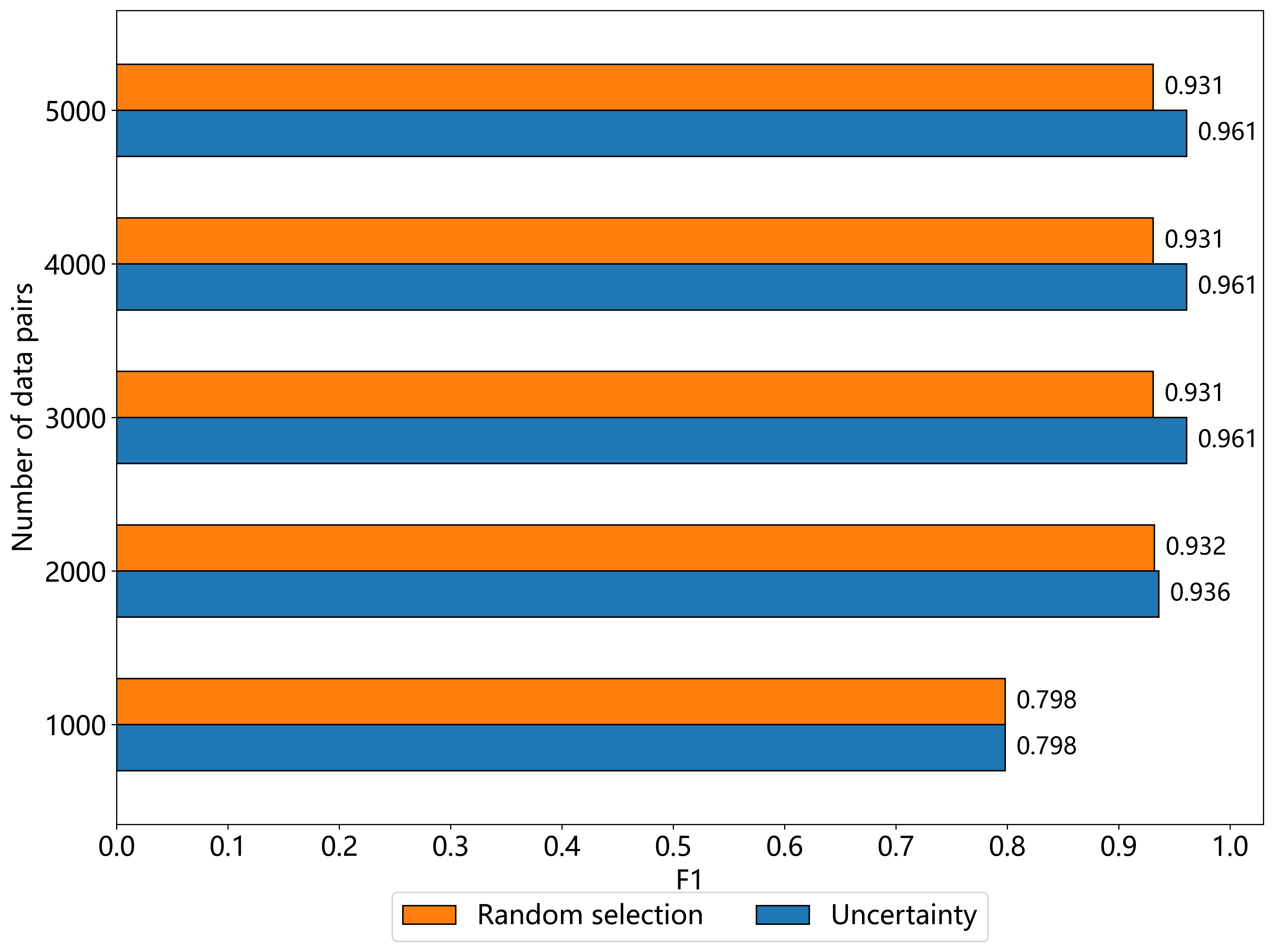}
	\end{center}
	\caption{The effect of different active learning selection strategies on their F1 on the GeographicSettlements dataset}
	\label{fig5}
\end{figure}

 Table \ref{tbl6} and Table \ref{tbl7} show the impact of different active learning selection strategies on F1 and Recall, and we compare the random selection strategy with the uncertainty strategy on the three datasets. As can be seen from the table, with the iterations of active learning, the F1 and Recall of the uncertainty selection strategy are higher than those of the random selection strategy. This is mainly due to the fact that the uncertainty-based selection strategy can select the most uncertain data of the model to be given to the experts for labelling in each round of active learning, and then the model learns from this part of the labelled data to improve its own accuracy. For example, as shown in the figure \ref{fig5}, which illustrates the effect of the two selection strategies on model F1 in the GeographicSettlements dataset. It can be seen that the uncertainty selection strategy can effectively improve the model accuracy, for example, after the third round of active learning, the uncertainty-based selection strategy is 3\% higher than the randomly selected selection strategy F1.

\par To sum up, PDDM-AL performs better than other models in duplicate data recognition, both for Precision, Recall and F1 rate. The main reasons are: firstly, the pre-trained model can learn the contextual information and semantics better than other methods; secondly, PDDM-AL injects domain knowledge, which can capture key information better than the simple pre-trained model, and also employs the R-Drop data enhancement strategy, which enables the model to maintain higher F1 and recall rates even with more dirty data, making the model more robust.

\section{Conclusion}
This study addresses the key challenges of data deduplication by proposing a pre-trained data deduplication model based on active learning. This is the first work to utilize active learning to address the problem of deduplication at the semantic level. The model is built on a pre-trained Transformer-based language model that has been fine-tuned to solve the deduplication problem as a sequence to classification task. Our proposed model, integrates the transformer with active learning into an end-to-end architecture to select the most valuable data for model training. In addition, we employ the R-Drop method to perform data augmentation on each round of labeled data to reduce the cost of manual labeling in big data and improve the model's performance with a small amount of labeled data. Experimental results demonstrate that PDDM-AL outperforms other benchmark models, showing the effectiveness of active learning in data deduplication. In the future, we will continue to explore different selection strategies and combinations of different models to further improve the accuracy of the data deduplication model.

\section*{Acknowledgments}

This work was supported by the National Natural Science Foundation of China (Nos. 62276215, 62272398, 62176221, 61976247).

\bibliographystyle{cas-model2-names}

\bibliography{mybibfile}



\end{document}